\documentclass[runningheads]{llncs}

\usepackage[T1]{fontenc}

\usepackage{amsmath}
\usepackage{amssymb}
\usepackage{amsfonts}
\usepackage{mathtools}

\usepackage{graphicx}
\usepackage{booktabs}
\usepackage{tabularx}
\usepackage{array}
\usepackage{multirow}
\usepackage{rotating}

\usepackage[dvipsnames]{xcolor}

\usepackage[ruled,vlined]{algorithm2e}

\usepackage{pgfplots}
\pgfplotsset{compat=1.18}
\usepgfplotslibrary{groupplots}
\usetikzlibrary{plotmarks}

\usepackage{subcaption}
\usepackage{placeins}    % provides \FloatBarrier

\usepackage{pifont}
\usepackage{comment}

\usepackage{hyperref}
\hypersetup{
    colorlinks=true,
    breaklinks=true,
    linkcolor=red,
    filecolor=magenta,
    urlcolor=blue,
    citecolor=blue
}
\newcommand{\best}[1]{\textbf{#1}}
\newcommand{\secondbest}[1]{\underline{#1}}

\begin{document}
\title{FastTab: A Fast Table Recognizer with a Tiny Recursive Module and 1D Transformers}
\titlerunning{FastTab: A Fast Table Recognizer with a TRM and 1D Transformers}
% If the paper title is too long for the running head, you can set
% an abbreviated paper title here
%
\author{Laziz Hamdi\inst{1,2}, 
Amine Tamasna\inst{2}, 
Pascal Boisson\inst{2} \and Thierry Paquet\inst{1}}
%
% \authorrunning{L.Hamdi et al.}
% First names are abbreviated in the running head.
% If there are more than two authors, 'et al.' is used.

\institute{LITIS, Rouen Normandy, France \and Malakoff Humanis, Paris, France} 
\maketitle              % typeset the header of the contribution
\begin{abstract}
% ============================================================
% 1) Abstract + keywords (currently placeholder / missing content)
% ============================================================
% In the current PDF, the abstract is essentially empty and the keywords are placeholders. :contentReference[oaicite:1]{index=1}
% You can drop-in something like this (adapt numbers if you change them later).
Table structure recognition (TSR) requires both table-level coherence (row/column counts, headers, spanning cells) and precise separator localization. We introduce \textsc{FastTab}, a grid-centric TSR model that avoids autoregressive HTML decoding by combining (i) a lightweight Tiny Recursive Module (TRM) for global reasoning and (ii) axial 1D Transformer encoders that capture long-range dependencies along rows and columns. The model predicts row/column counts, header rows, and separators to construct a grid, then infers rowspan/colspan using ROI-aligned cell features. Across four benchmarks (PubTabNet, FinTabNet, PubTables-1M, and SciTSR), \textsc{FastTab} achieves competitive structure recovery performance  while operating at low-latency inference. We further study robustness under pixel-level anonymisation and show an extension to curved separators for camera-captured documents. The source code will be made publicly available at \url{https://github.com/hamdilaziz/FastTab}.

\end{abstract}
\section{Introduction}
\label{sec:intro}

Tables are pervasive in professional and scientific documents because they aggregate relational information into a 2D structured visual representation. Recovering this information automatically from images remains a central problem in document intelligence: a correct extraction must preserve the logical organization of the table (rows, columns, header regions, and merged cells) and maintain geometric fidelity (cell extents consistent with the layout). The difficulty increases in real documents that contain borderless designs, dense multi-row headers, heterogeneous typography, and imperfect alignment, where visual cues are incomplete and multiple structural interpretations are possible \cite{YU2024128154,huang2024detection}.

Table structure recognition (TSR) has evolved from rule-based pipelines toward Machine Learning based models that infer structure from data. Early Machine Learning based systems often decomposed TSR into component prediction and relational reconstruction, i.e. by detecting table elements and then inferring row/column relations using graph-based reasoning \cite{Schreiber2017DeepDeSRTDL,paliwal2019tablenet,Xue2021TGRNet,Liu2022NCGM,Qiao2021LGPMA}. Some other approaches introduce a split-and-merge paradigm to predict row and column separators to form an initial grid, and then infer merged cells through a merging stage \cite{tensmeyer2019deep,zhang2022split,guo2208trust,lin2022tsrformer}. More recently, image-to-sequence models have treated TSR as a sequence generation problem, enabled by large-scale datasets that exploit HTML markup of the tables as the output structure to predict (used as the GT during supervised training). Such approaches have also introduced new structure-aware evaluation metrics such as TEDS \cite{zhong2020image} (Tree-Edit-Distance-based Similarity). Such models generate HTML (or related structured languages) using encoder-decoder-based architectures \cite{zhong2020image,nassar2022tableformer,huang2023improving}. These three lines of work provide complementary perspectives: component-based methods emphasize local evidence and explicit relations, split-and-merge methods emphasize explicit grid construction, and generative-based methods emphasize a direct mapping to structured markup.

A recurring difficulty in these paradigms is reconciling global structural coherence with local spatial precision. Separator localization and borderless layout cues depend on fine-grained information, axis-aligned evidence, while decisions such as row/column counts and header extent relate to table-level context. FastTab is designed around this tension. Rather than relying on a long autoregressive Transformer decoder \cite{zhong2020image,nassar2022tableformer,huang2023improving}, or on multi-stage relational reconstruction \cite{Schreiber2017DeepDeSRTDL,Xue2021TGRNet,Liu2022NCGM}, FastTab follows a grid-centric formulation related in spirit to split-and-merge pipelines \cite{tensmeyer2019deep,zhang2022split,guo2208trust,lin2022tsrformer}, but differs in how global context and axial dependencies are modeled. Specifically, FastTab combines (i) a Tiny Recursive Module~\cite{jolicoeur2025less} that iteratively refines a global latent vector from pooled visual features to support table-level decisions, and (ii) axial 1D Transformer encoders that process row-wise and column-wise sequences derived from the 2D feature map, providing long-range context along each principal axis with a computation that grows linearly with the axis length. Conditioned on the predicted grid, merged cells are inferred by ROI-aligned pooling over the feature map followed by rowspan/colspan classification, supervised at anchor positions to avoid ambiguous targets inside covered regions. 

This paper makes three contributions.
\begin{itemize}
    \item Introduces \textsc{FastTab}, an efficient grid-centric TSR model that combines a Tiny Recursive Module for global context with axial 1D Transformers for long-range row/column reasoning
    \item Reports structure recovery performance and runtime on four established TSR benchmarks, showing that \textsc{FastTab} is competitive while operating at low latency.
    \item Analyzes key design choices through ablations and evaluates robustness under two practical settings: pixel-level anonymisation and curved separators.
\end{itemize}

\begin{figure*}[!t]
\centering
\includegraphics[width=\linewidth]{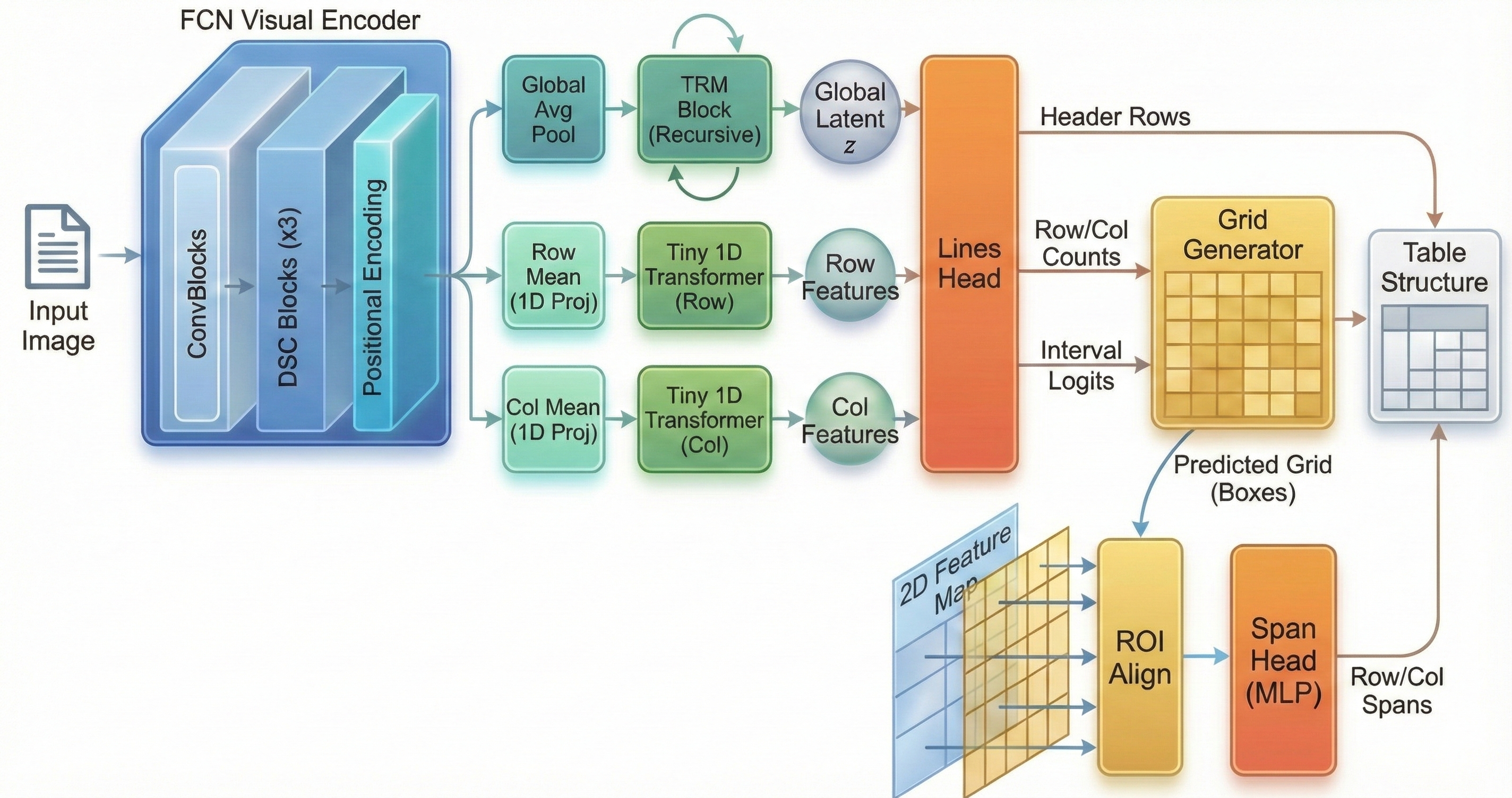}
\caption{Overview of FastTab. An FCN encoder extracts a 2D feature map, which is summarized by a global TRM branch and two axial (row/column) 1D Transformer branches. A Lines Head predicts row/column counts, header rows, and separator/interval logits to build the grid. ROI Align then pools cell features for a Span Head that predicts rowspan/colspan, yielding the final table structure.}
\label{framework}
\end{figure*}

\section{Related Work}
\label{sec:related_work}

Table structure recognition (TSR) has been approached through several modeling paradigms that differ in the intermediate representations they predict and in how they enforce consistency between table-level structure and local geometry. In this section, we briefly review the most relevant lines of work and position FastTab with respect to them.

\subsection{Component-based and relation-based TSR}
Early machine-learning-based TSR systems commonly decomposed the problem into visual components prediction, followed by an explicit reconstruction stage. DeepDeSRT integrates table detection and structure recognition in an end-to-end framework tailored to document images \cite{Schreiber2017DeepDeSRTDL}. TableNet predicts table and column regions using a shared backbone and then applies rule-based steps for row extraction \cite{paliwal2019tablenet}. Beyond segmentation-oriented decompositions, several methods formulate TSR as relational reasoning over detected cells or text regions. GraphTSR models TSR in PDF documents by predicting the relations among table cells and introduced the SciTSR dataset to evaluate spanning-cell structures \cite{chi2019complicated}. TGRNet similarly emphasizes graph-based reconstruction by predicting cell-level relationships to recover table topology \cite{Xue2021TGRNet}. NCGM studies heterogeneous tables and models interactions between modalities using collaborative graph reasoning \cite{Liu2022NCGM}. LGPMA improves cell boundary localization through local and global pyramid mask alignment and proposes a structure recovery pipeline that addresses issues such as empty cell division \cite{Qiao2021LGPMA}. These methods make relational constraints explicit, but final performance is often dependent on the quality of the intermediate detections and the robustness of the subsequent reconstruction.

\subsection{Split-and-merge and separator-driven models}
A second line of work constructs a table grid by predicting row and column separators, and then infers merged cells by merging adjacent grid regions. Deep Splitting and Merging learns a splitting stage to estimate the base grid and a merging stage to recover spanning cells \cite{tensmeyer2019deep}. SEM formalizes this approach into a splitter-embedder-merger design, where visual and textual information are fused at the grid level before learning merge decisions \cite{zhang2022split}. SEMv2 improves the 'split' stage by treating separation lines as instances and using conditional convolution for instance-level discrimination \cite{zhang2023semv2}. TRUST proposes a transformer-based splitting module for multi-oriented separators and a vertex-based merging module for spanning-cell recovery, and reports real-time inference on PubTabNet in its experimental setting \cite{guo2208trust}. TSRFormer formulates separator prediction as a regression problem and uses a DETR-style design to improve localization while keeping computation manageable \cite{lin2022tsrformer}. This family of approaches is closely related to FastTab in that it explicitly parameterizes the grid and separates boundary estimation from span inference, but existing methods typically rely on specialized separator prediction modules or multi-stage designs to maintain both accuracy and robustness under distortions.

\subsection{End-to-end sequence generation}
A third family of approaches directly produces a structured representation, most commonly by introducing a generative decoder that produces the desired sequence of tokens. PubTabNet popularized image-to-sequence table recognition by introducing large-scale HTML supervision, an encoder-dual-decoder architecture, and the TEDS (Tree-Edit Distance Similarity) metric for structure-aware evaluation \cite{zhong2020image}. TableFormer improves the PubTabNet-style baseline by introducing a transformer-based design that explicitly models table structure and integrates an object-detection decoder for table cells \cite{nassar2022tableformer}. Several works focus on strengthening the coupling between logical structure tokens and geometric grounding. VAST strengthens the link between structure tokens and geometry by predicting cell coordinates with a dedicated coordinate decoder module and by adding an alignment loss during training \cite{huang2023improving}. OTSL proposes a compact structure language to reduce the sequence length and vocabulary size relative to HTML, improving runtime characteristics for sequence-based models \cite{lysak2023optimized}. In contrast to token generation, LORE frames TSR as logical location regression and jointly regresses logical and spatial locations of cells \cite{xing2023lore}. TFLOP reformulates region matching as a layout-pointer problem to reduce misalignment between text regions and structure tags, and introduces span-aware supervision to support complex tables \cite{khang2025tflop}. Generative models provide a direct path to structured output generation, but they inherit the computational cost of autoregressive decoding, and the need to maintain global syntactic and structural consistency along the sequence. In contrast, regression-based models emphasize geometric grounding and may require careful design to reliably represent complex spans and headers.

\subsection{Positioning of FastTab}
FastTab is most closely aligned with separator-driven grid construction approaches in that it explicitly predicts row/column counts, boundary locations, and merged-cell spans. Its main distinguishing feature is that these predictions are obtained with a computation-efficient backbone: a lightweight recursive global reasoning module and two axial 1D Transformer encoders operating on row-wise and column-wise sequences obtained by pooling a 2D feature map. By restricting self-attention to 1D axes and avoiding a full autoregressive decoder \cite{zhong2020image,nassar2022tableformer,huang2023improving}, FastTab reduces decoding overhead and enables high-throughput inference, while still modeling long-range dependencies required for consistent separators and spans. At the same time, the output remains an explicit grid parameterization, comparable in spirit to split-and-merge pipelines \cite{tensmeyer2019deep,zhang2022split,guo2208trust,lin2022tsrformer}, but with fewer sequential steps and a simpler inference path.
\section{Methodology}
\label{sec:method}

FastTab predicts a table grid and merged cells directly, without generating HTML or other markup. Given an input image, the model first extracts a 2D feature map with an FCN visual encoder. It then predicts (i) the number of rows and columns, the number of header rows, and the row/column separators using the Lines Head (conditioned on a global latent vector $z$ and axial row/column features), and (ii) the rowspan/colspan of each logical cell using ROI Align and a Span Head (MLP).

\paragraph{\textbf{Problem definition}}
Let $\mathbf{I}\in\mathbb{R}^{H\times W\times 3}$ be an image that contains a single table (or a cropped region that tightly contains the table). The table has $R$ rows and $C$ columns. We represent the grid by its horizontal and vertical boundary coordinates,
\begin{equation}
\mathbf{y}=(y_0,\ldots,y_R),\qquad \mathbf{x}=(x_0,\ldots,x_C),
\label{eq:bounds_def}
\end{equation}
where each coordinate is expressed in normalized table coordinates. Hence $y=0$ and $y=1$ denote the top and bottom of the table region, while $x=0$ and $x=1$ denote the left and right boundaries. The boundaries are ordered along each axis,
\begin{equation}
0=y_0 \le y_1 \le \cdots \le y_R=1,\qquad 0=x_0 \le x_1 \le \cdots \le x_C=1,
\label{eq:bounds_order}
\end{equation}
so that each row and column corresponds to a valid interval. Merged cells are represented by rowspan and colspan values $(\mathrm{rs}_{r,c},\mathrm{cs}_{r,c})$, defined only at the top-left position (anchor) of each logical cell. We also predict the number of header rows $H_{\mathrm{hdr}}\in\{0,\ldots,R\}$, interpreted as the number of top grid rows belonging to the header region.

\paragraph{\textbf{Image encoder}}
The image is mapped to a high-resolution 2D feature map using a fully-convolutional encoder,
\begin{equation}
\mathbf{F}=E(\mathbf{I})\in\mathbb{R}^{d_{\mathrm{model}}\times H_f\times W_f},
\label{eq:fmap}
\end{equation}
where $d_{\mathrm{model}}$ is the channel dimension and $(H_f,W_f)$ is the spatial resolution after downsampling. FastTab uses an anisotropic stride schedule that yields $H_f=H/16$ and $W_f=W/8$ (four downsampling stages along height, but only three along width), which preserves finer horizontal sampling for column delineation. The encoder outputs $d_{\mathrm{model}}=1024$ channels.

\paragraph{\textbf{Tiny Recursive Module}}
Row/column counts and header extent benefit from table-level context. We first compute a global descriptor by average pooling,
\begin{equation}
\mathbf{g}=\mathrm{GAP}(\mathbf{F})\in\mathbb{R}^{d_{\mathrm{model}}}.
\label{eq:gap}
\end{equation}
We then refine a latent vector $\mathbf{z}\in\mathbb{R}^{d_z}$ over $T$ residual steps using a Tiny Recursive Module (TRM):
\begin{equation}
\mathbf{z}^{(0)}=\mathbf{z}_0,\qquad
\mathbf{z}^{(t+1)}=\mathbf{z}^{(t)}+\psi\!\left([\mathrm{LN}(\mathbf{z}^{(t)}),\mathbf{g}]\right),\qquad t=0,\ldots,T-1,
\label{eq:trm}
\end{equation}
where $\mathbf{z}_0$ is a learned initialization vector and $\mathrm{LN}$ denotes layer normalization. The update function $\psi$ is a lightweight two-layer MLP that maps $\mathbb{R}^{d_z+d_{\mathrm{model}}}\rightarrow \mathbb{R}^{d_z}$:
\begin{equation}
\psi(\mathbf{u})=\mathbf{W}_{\mathrm{out}}\;\mathrm{GELU}(\mathbf{W}_{\mathrm{in}}\mathbf{u}),
\label{eq:trm_mlp}
\end{equation}
with $\mathbf{W}_{\mathrm{in}}\in\mathbb{R}^{d_z\times(d_z+d_{\mathrm{model}})}$ and $\mathbf{W}_{\mathrm{out}}\in\mathbb{R}^{d_z\times d_z}$, $T=6$ and $d_z=1024$. The final vector $\mathbf{z}^{(T)}$ conditions the Lines Head.

\begin{figure}[!t]
\centering
\includegraphics[width=0.95\linewidth]{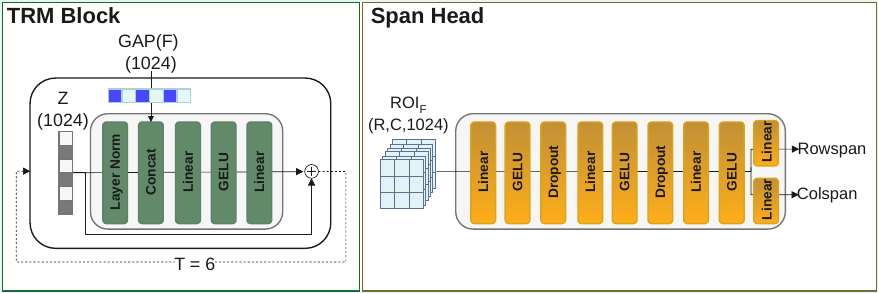}
\caption{Architecture details of the Span Head and Tiny Recursive Module.}
\label{TRM_SPAN}
\end{figure}
\paragraph{\textbf{Lines Head with a Tiny 1D Transformer}}
To predict separators, FastTab reduces the 2D feature map into two axial sequences by averaging along the orthogonal dimension:
\begin{equation}
\mathbf{S}^{(r)}=\mathrm{mean}_{w}(\mathbf{F})\in\mathbb{R}^{d_{\mathrm{model}}\times H_f},
\qquad
\mathbf{S}^{(c)}=\mathrm{mean}_{h}(\mathbf{F})\in\mathbb{R}^{d_{\mathrm{model}}\times W_f}.
\label{eq:rowcol_seq}
\end{equation}
Each sequence is projected to $d_{\mathrm{seq}}=256$ using a $1\times 1$ Conv1D and processed along its length by a tiny Transformer encoder with learned 1D positional embeddings and a PreNorm configuration ($L=2$ layers, $A=4$ attention heads, $d_{\mathrm{ff}}=512$, dropout $0.1$, GELU). A lightweight depthwise Conv1D (kernel size $3$) followed by a pointwise $1\times 1$ Conv1D and GELU restores local smoothness, producing contextual axial features $\widetilde{\mathbf{S}}^{(r)}\in\mathbb{R}^{d_{\mathrm{seq}}\times H_f}$ and $\widetilde{\mathbf{S}}^{(c)}\in\mathbb{R}^{d_{\mathrm{seq}}\times W_f}$. The head combines global summaries of these features with the TRM latent $\mathbf{z}^{(T)}$ to predict row/column counts $(R,C)$ and header rows $H_{\mathrm{hdr}}$.

Separator locations are parameterized by normalized interval lengths. For each axis, the head outputs fixed-length logits $\mathbf{o}^{(r)}\in\mathbb{R}^{R_{\max}+1}$ and $\mathbf{o}^{(c)}\in\mathbb{R}^{C_{\max}+1}$ via adaptive pooling followed by a shared linear projection. Keeping the first $R$ (resp.\ $C$) logits and applying a softmax yields length vectors $\mathbf{p}^{(r)}$ (resp.\ $\mathbf{p}^{(c)}$), which are converted into ordered boundaries by cumulative summation:
\begin{equation}
y_0=0,\qquad y_k=\sum_{i=1}^{k} p^{(r)}_i,\ k=1,\ldots,R,
\qquad
x_0=0,\qquad x_k=\sum_{i=1}^{k} p^{(c)}_i,\ k=1,\ldots,C.
\label{eq:cumsum_rows_cols}
\end{equation}
This construction enforces monotonicity and ensures that the final boundary equals $1$; the resulting grid is passed to the Grid Generator.

\paragraph{\textbf{Spanning Head}}
Once $(\mathbf{y},\mathbf{x})$ and $(R,C)$ are available, the Grid Generator defines one rectangular region per grid cell. ROI Align pools one feature vector per region from $\mathbf{F}$ (output size $1\times 1$), yielding a tensor
\begin{equation}
\mathbf{U}\in\mathbb{R}^{R\times C\times d_{\mathrm{model}}}.
\label{eq:roi_tensor}
\end{equation}
A Span Head MLP maps each pooled feature to rowspan and colspan logits. The MLP uses three linear stages,
\begin{equation}
d_{\mathrm{model}}(=1024)\ \rightarrow\ 512\ \rightarrow\ 512\ \rightarrow\ 256,
\end{equation}
with GELU activations and dropout ($p=0.1$) after the first two stages, and two linear classifiers output
$\mathrm{rs}_{r,c}\in\{1,\ldots,RS_{\max}\}$ and $\mathrm{cs}_{r,c}\in\{1,\ldots,CS_{\max}\}$. During training, span supervision is applied only at anchor positions (top-left corners of logical cells), while covered positions inside merged regions are ignored.

\paragraph{\textbf{Training}}
\label{training}
FastTab is trained end-to-end with four loss terms. Row and column counts are supervised with cross-entropy losses, and the number of header rows is supervised with a cross-entropy loss over $\{0,\ldots,R_{\max}\}$. Separator positions are supervised via the cumulative boundary coordinates using a mean-squared error loss, which directly penalizes boundary displacement. Rowspan and colspan are supervised with cross-entropy losses computed only at anchor locations (top-left corners of logical cells); anchor positions corresponding to merged cells can be up-weighted to mitigate class imbalance.

The span head relies on ROI regions derived from the predicted grid. Early in training, inaccurate separators can yield unstable ROI pooling, so we use a teacher-forcing schedule for ROI construction: a subset of samples uses ground-truth separators for ROI generation, and this fraction is progressively annealed to $20\%$. When ground-truth separators are used, we add small random perturbations to encourage robustness to boundary errors at inference.
\section{Experiments}
\label{sec:exp}

\begin{figure}[!t]
\centering
\includegraphics[width=\linewidth]{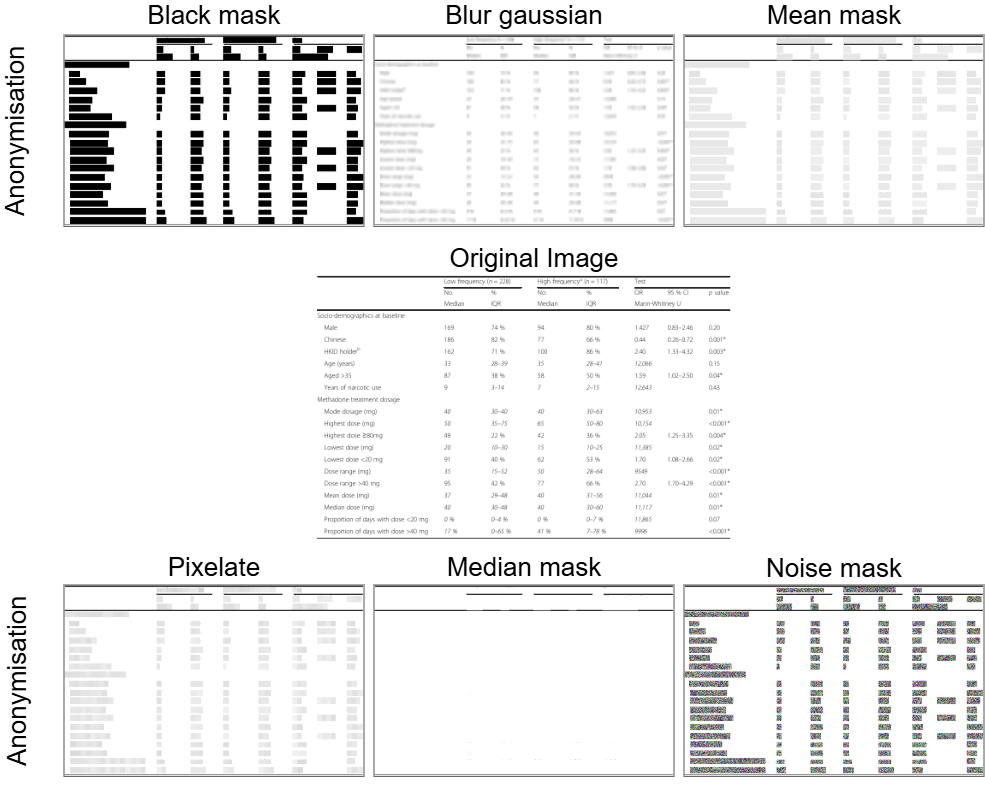}
\caption{PubTabNet example (\texttt{PMC4608158\_004\_00.png}) and its anonymised variants.}
\end{figure}

\paragraph{Datasets}
We evaluate FastTab for table structure recognition (TSR) on four established benchmarks spanning scientific and financial domains. PubTabNet~\cite{zhong2020image} contains table images from scientific articles paired with HTML annotations, and is widely used to assess image-based structure recovery under diverse visual styles. FinTabNet~\cite{zheng2021global} targets financial tables extracted from reports, with structure annotations derived from PDF sources, complementing PubTabNet in both domain and layout characteristics. PubTables-1M~\cite{Smock2022PubTables1M} is a large-scale corpus of scientific tables with cell-level bounding-box annotations; following the official protocol, it is split at the document level (80/10/10) into 758{,}849/94{,}959/93{,}834 tables. Finally, SciTSR~\cite{chi2019complicated} contains tables extracted from scientific PDFs with structure labels; the standard split includes 12{,}000 training samples and 3{,}000 test samples.

We set dataset-specific caps $(R_{\max},C_{\max},RS_{\max},CS_{\max})$ to $(50,30,10,10)$ (PubTabNet), $(85,30,10,25)$ (FinTabNet), $(85,52,55,40)$ (PubTables-1M), and $(60,66,17,17)$ (SciTSR), with batch sizes $(16,10,10,8)$ respectively. Optimization uses AdamW with cosine annealing (initial learning rate $10^{-5}$).

\paragraph{Metrics}
We report three complementary metrics following benchmark conventions. On PubTabNet and FinTabNet, we use S-TEDS, i.e., the structure-only variant of Tree-Edit-Distance-based Similarity (TEDS) that removes cell text before computing similarity in HTML-tree space~\cite{zhong2020image}; this quantity is also reported as \emph{TEDS-Struct} in prior work. On PubTables-1M, we report GriTS$_\text{Top}$~\cite{Smock2022PubTables1M}, which evaluates recovered grid topology (rows/columns and spanning structure) by comparing predicted and ground-truth grids. On SciTSR, we follow the standard protocol and report CAR F1, the F1-measure over cell adjacency relations~\cite{chi2019complicated}.

\paragraph{Runtime protocol and comparability}
For FastTab, FPS is measured on a single NVIDIA A100 GPU with batch size $1$ and without resizing. The reported time includes image transfer, a forward pass, and construction of the predicted HTML grid, and excludes any external OCR. For prior work, we quote FPS from the corresponding papers when available; since protocols differ (hardware, resizing, post-processing, and text-related components), FPS values are indicative rather than strictly comparable.

\paragraph{Results on \textit{PubTabNet and FinTabNet}}
Table~\ref{tab:tsr_tcr_results} summarizes accuracy--speed trade-offs on PubTabNet and FinTabNet. FastTab operates in the real-time regime while achieving competitive structure accuracy. On a CPU-only environment (no GPU), FastTab still runs at 5.24 FPS on PubTabNet and 4.66 FPS on FinTabNet, supporting its suitability for resource-constrained deployments.

\begin{figure}[!t]
\centering
\includegraphics[width=0.95\linewidth]{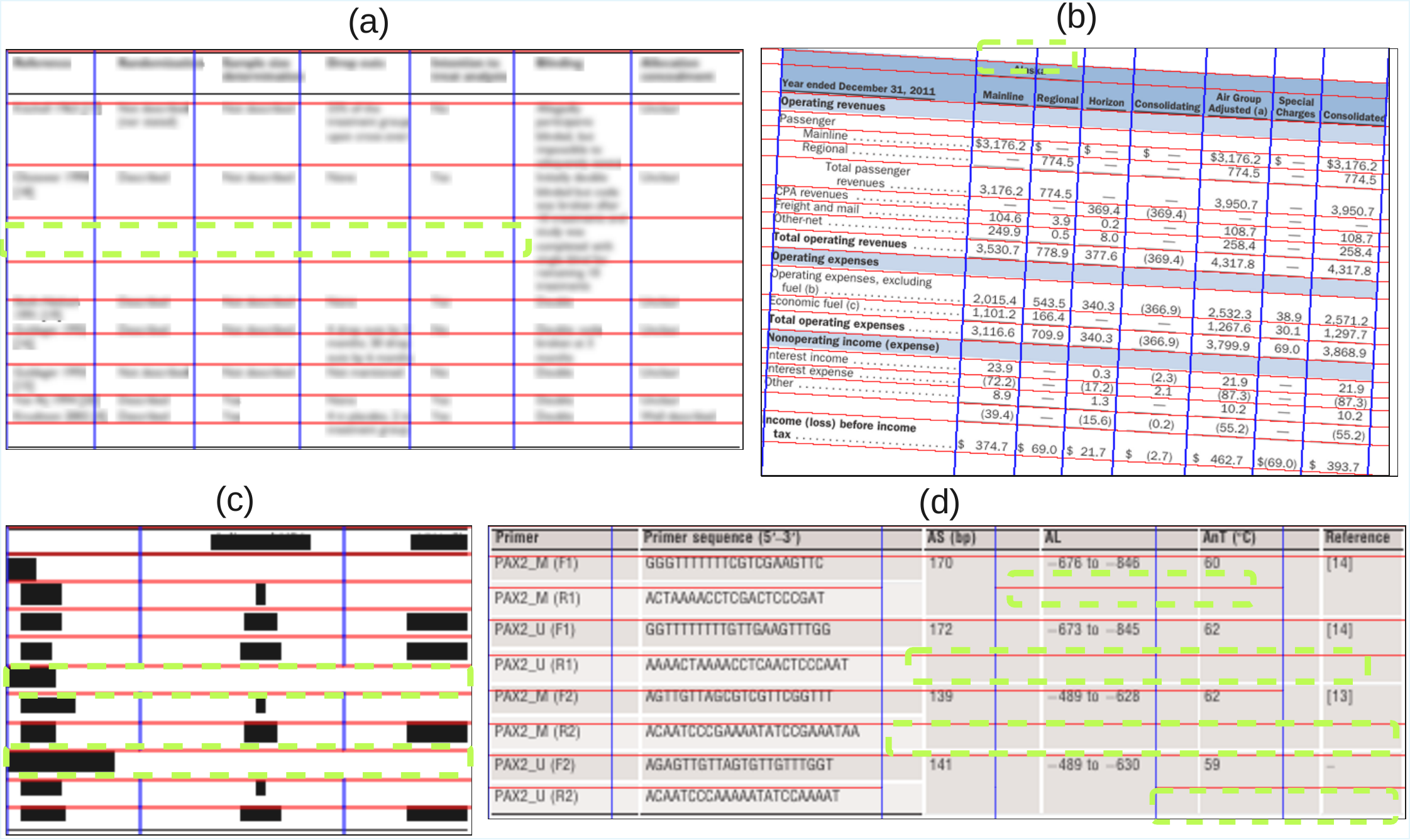}
\caption{Representative predictions. Red/blue: row/column separators; dashed green: errors. Examples include black-mask, curved, Gaussian-blur, and original samples from PubTabNet/FinTabNet.}
\label{representatives_samples}
\end{figure}

\begin{table*}[t]
  \centering
  \footnotesize
  \caption{TSR comparison on PubTabNet and FinTabNet test sets.}
  \label{tab:tsr_tcr_results}
  \setlength{\tabcolsep}{4pt}
  \renewcommand{\arraystretch}{0.95}

  % \extracolsep{\fill} distributes remaining space evenly between columns
  \begin{tabular*}{\textwidth}{@{\extracolsep{\fill}}p{0.22\textwidth}cccc@{}}
    \toprule
    \multirow{2}{*}{\textbf{Method}} &
    \multicolumn{2}{c}{PubTabNet} &
    \multicolumn{2}{c}{FinTabNet}\\
    \cmidrule(lr){2-3}\cmidrule(lr){4-5}
    & \textbf{S-TEDS} & \textbf{FPS} & \textbf{S-TEDS} & \textbf{FPS}\\
    \midrule
    EDD~\cite{zhong2020image}           & 89.9  & 1.0   & 90.6  & --   \\
    SEMv2~\cite{zhang2023semv2}         & \secondbest{97.5} & --    & 92.8  & --  \\
    TSRFormer~\cite{lin2022tsrformer}   & \secondbest{97.5} & --    & --    & --   \\
    TRUST~\cite{guo2208trust}           & 97.1  & 10.0  & --    & --   \\
    RTSR~\cite{nguyen2024rtsr}          & 95.7  & \secondbest{43.2} & 97.2 & \secondbest{38.0} \\
    VAST~\cite{huang2023improving}      & 97.2  & 1.38  & \secondbest{98.6} & 1.38 \\
    TABLET~\cite{hou2025tablet}         & \best{97.7} & --    & \best{98.7} & 18.0 \\
    OmniParser~\cite{wan2024omniparser} & 90.5  & 1.3   & 91.6  & 1.3  \\
    GridFormer~\cite{lyu2023gridformer} & 97.0  & --    & \secondbest{98.6} & --   \\
    LGPMA~\cite{Qiao2021LGPMA}          & 96.7  & --    & --    & --   \\
    \midrule
    FastTab                             & 96.8  & \best{44.89} & 98.2 & \best{41.79}\\
    \bottomrule
  \end{tabular*}

% \vspace{2pt}
% \parbox{\textwidth}{\footnotesize
% \emph{Notes.} FPS values are taken from the original papers and are not directly comparable due to differences in hardware, resizing, and post-processing. When specified: TRUST uses A100 with $640{\times}640$ inputs; RTSR reports RTX~3060 with long-side resizing; VAST resizes to $608{\times}608$ and uses autoregressive decoding; TABLET includes resizing and post-processing.}
\end{table*}

\paragraph{Results on \textit{PubTables-1M}.}
Table~\ref{tab:pubtables_results} reports TSR accuracy on PubTables-1M using GriTS$_\text{Top}$, which directly evaluates grid topology and spanning-cell recovery. FastTab achieves strong performance (98.27) while remaining highly efficient (39.50 FPS under our A100/bs=1 protocol). Since most prior work do not report PubTables-1M inference speed under a clearly specified and comparable setting, we primarily use PubTables-1M to position FastTab in terms of topology accuracy, and rely on PubTabNet/FinTabNet/SciTSR for a more informative runtime comparison.

\begin{table}[t]
  \centering
  \small
  \caption{TSR comparison on \textbf{PubTables-1M} (test). Metric is GriTS$_\text{Top}$ (\%). FPS is reported only when explicitly provided.}
  \label{tab:pubtables_results}
  \begingroup
  \fontsize{9}{10.5}\selectfont
  \setlength{\tabcolsep}{8pt}
  \renewcommand{\arraystretch}{0.95}
  \begin{tabularx}{0.65\linewidth}{p{0.3\textwidth} cc}
    \toprule
    \multicolumn{3}{c}{PubTables-1M} \\
    \midrule
    \textbf{Method} & \textbf{GriTS$_\text{Top}$} & \textbf{FPS} \\
    \midrule
    Faster R-CNN~\cite{Smock2022PubTables1M} & 86.16 & -- \\
    DETR~\cite{Smock2022PubTables1M}         & \secondbest{98.45} & -- \\
    DETR-NC~\cite{Smock2022PubTables1M}      & 97.62 & -- \\
    VAST~\cite{huang2023improving}           & \best{99.22} & -- \\
    \midrule
    FastTab                                  & 98.27 & 39.50 \\
    \bottomrule
  \end{tabularx}
  \endgroup
\end{table}

\paragraph{Results on \textit{SciTSR}}
Table~\ref{tab:scitsr_results} reports CAR F1 on SciTSR. FastTab matches the best reported accuracy (99.5 F1) and runs in real time. Its FPS is lower than RTSR in this comparison because we evaluate FastTab on the original image resolution, whereas RTSR reports timings with test-time resizing (long side fixed to 896). This highlights that reported FPS depends strongly on preprocessing choices and should be interpreted together with the stated evaluation protocol.

\begin{table*}[t]
  \centering
  \small
  \caption{Structure recognition comparison on \textbf{SciTSR} (test). Metric is CAR F1 (\%). FPS is reported when officially provided.}
  \label{tab:scitsr_results}
  \begingroup
  \fontsize{9}{10.5}\selectfont
  \setlength{\tabcolsep}{8pt}
  \renewcommand{\arraystretch}{0.95}
  \begin{tabularx}{0.7\linewidth}{p{0.4\textwidth} cc}
    \toprule
    \multicolumn{3}{c}{SciTSR} \\
    \midrule
    \textbf{Method} & \textbf{F1 score} & \textbf{FPS} \\
    \midrule
    TabStruct-Net~\cite{chi2019complicated}          & 92.0 & -- \\
    GraphTSR~\cite{chi2019complicated}              & 95.3 & -- \\
    LGPMA~\cite{Qiao2021LGPMA}                      & 98.8 & -- \\
    FLAG-Net~\cite{liu2021show}                      & \best{99.5} & -- \\
    SEM~\cite{zhang2022split}                       & 97.1 & 1.94 \\
    SEMv2~\cite{zhang2023semv2}                     & 99.3 & 7.3 \\
    TSRFormer w/ DQ-DETR~\cite{lin2022tsrformer}    & 98.9 & 4.2 \\
    LORE++~\cite{long2025lore++}                    & --   & 2.3 \\
    RTSR~\cite{nguyen2024rtsr}                      & \secondbest{99.4} & \best{45.8} \\
    \midrule
    FastTab                                         & \best{99.5} & \secondbest{35.51} \\
    \bottomrule
  \end{tabularx}
  \endgroup
  \vspace{2pt}
  \parbox{\textwidth}{\footnotesize\emph{Note:} RTSR reports long-side resizing to 896 and inference on an RTX~3060; SEM/SEMv2 report experiments on a Tesla V100 (32GB) with batch size 8.}
\end{table*}

\subsection{TSR with Pixel-Level Anonymisation}
\label{subsec:tsr_anonym}

Industrial deployments often require table images to be anonymised before inference. To approximate this setting, we construct an anonymised variant of PubTabNet~\cite{zhong2020image} using its text bounding boxes: for each image, we sample one anonymisation method and apply it to all text boxes, removing readable content while preserving the table layout. At evaluation time, we run FastTab with a fixed anonymisation method per setting. Table~\ref{tab:anonym_results} reports S-TEDS (\%) on the simple/complex subsets and overall; \emph{Baseline} corresponds to the original (non-anonymised) images.

Results follow the expected trend: methods that preserve low-frequency layout cues are least harmful, while strong texture replacement and high-frequency corruption degrade structure recovery. Gaussian blur and pixelation cause moderate overall drops (97.4$\rightarrow$96.7 and 97.4$\rightarrow$96.2), with blur essentially preserving the complex subset (94.5$\rightarrow$94.7) and pixelation yielding a larger complex drop (94.5$\rightarrow$92.9). Black masking and mean fill reduce S-TEDS to 95.1 and 90.7, while additive noise is also damaging (86.1). Median fill is the most disruptive (77.3), likely because it homogenizes large text-box regions and suppresses weak structural cues. Qualitative examples are shown in Fig.~\ref{representatives_samples}.

\begin{table}[t]
  \centering
  \small
  \caption{FastTab on \textbf{Anonymised PubTabNet} (validation). Metric is S-TEDS (\%) reported on the simple/complex subsets and overall. \emph{Baseline} denotes evaluation without anonymisation.}
  \label{tab:anonym_results}
  \begingroup
  \fontsize{9}{10.5}\selectfont
  \setlength{\tabcolsep}{7pt}
  \renewcommand{\arraystretch}{0.95}
  \begin{tabularx}{0.75\linewidth}{p{0.27\textwidth} c c c}
    \toprule
    \textbf{Anonymisation} & \textbf{Simple} & \textbf{Complex} & \textbf{Overall} \\
    \midrule
    Baseline             & 99.1 & 94.5 & 97.4 \\
    Black                & 97.3 & 91.3 & 95.1 \\
    Median               & 82.1 & 73.0 & 77.3 \\
    Gaussian blur        & 99.0 & 94.7 & 96.7 \\
    Pixelation           & 98.4 & 92.9 & 96.2 \\
    Mean                 & 94.2 & 90.0 & 90.7 \\
    Noise                & 89.9 & 82.0 & 86.1 \\
    \bottomrule
  \end{tabularx}
  \endgroup
\end{table}

\subsection{Curved separators under rotation}
\label{subsec:tsr_curved}

To improve tolerance to geometric perturbations (e.g., mild warping or in-plane rotation), we extend FastTab’s separator representation from straight boundaries to \emph{curved} boundaries. For each row (resp. column) boundary, the model predicts a polyline with $K{=}128$ uniformly sampled points, parameterized as $y_i(t_k)$ (resp. $x_j(t_k)$) in normalized table coordinates, with $t_k \in [0,1]$. Concretely, we keep the standard interval-based decoding to obtain a straight baseline grid, and predict a bounded residual offset for each boundary and each sample point; a smoothness penalty encourages locally regular curves, while a non-crossing regularizer discourages violations of the ordering constraint between adjacent boundaries.

\begin{figure}[t]
  \centering
  \begin{tikzpicture}
    \begin{axis}[
      width=0.95\linewidth,
      height=0.42\linewidth,
      xlabel={Rotation interval (degrees)},
      ylabel={S-TEDS (\%)},
      xmin=-1, xmax=31,
      ymin=92.8, ymax=100.,
      xtick={0,5,10,15,20,25,30},
      % IMPORTANT: each label must be wrapped in {...} because labels contain commas
      xticklabels={{$0^\circ$},{\ensuremath{[-5,5]}},{\ensuremath{[-10,10]}},
                   {\ensuremath{[-15,15]}},{\ensuremath{[-20,20]}},
                   {\ensuremath{[-25,25]}},{\ensuremath{[-30,30]}}},
      x tick label style={rotate=20, anchor=east},
      grid=major,
      major grid style={opacity=0.15},
      tick align=outside,
      tick label style={font=\small},
      label style={font=\small},
      legend style={font=\small, draw=none, fill=none},
      legend pos=south west,
    ]

      \addplot+[
        thick,
        mark=*,
        mark size=2.0pt,
        error bars/.cd,
          y dir=both,
          y explicit,
          error bar style={line width=0.9pt},
          error mark options={rotate=90, mark size=2.2pt, line width=0.9pt},
      ] coordinates {
        (0,  98.2) +- (0, 0.0)
        (5,  97.5) +- (0, 0.9)
        (10, 96.2) +- (0, 1.1)
        (15, 95.5) +- (0, 1.3)
        (20, 95.3) +- (0, 1.3)
        (25, 94.7) +- (0, 1.5)
        (30, 93.9) +- (0, 1.7)
      };
      \addlegendentry{Mean $\pm$ 1 std}

      \addplot+[dashed, thick] coordinates {(-1,98.2) (31,98.2)};
      \addlegendentry{Baseline (no rotation)}

      \node[font=\small, anchor=south east] at (axis cs:30.8,98.2) {98.2};

    \end{axis}
  \end{tikzpicture}
  \caption{\normalsize \textbf{FinTabNet rotation robustness.} Mean S-TEDS (\%) $\pm$ one standard deviation under random in-plane rotations sampled uniformly from $[-\alpha,\alpha]$. The dashed line indicates the no-rotation baseline (98.2).}
  \label{fig:curved_ftn_rotation}
\end{figure}
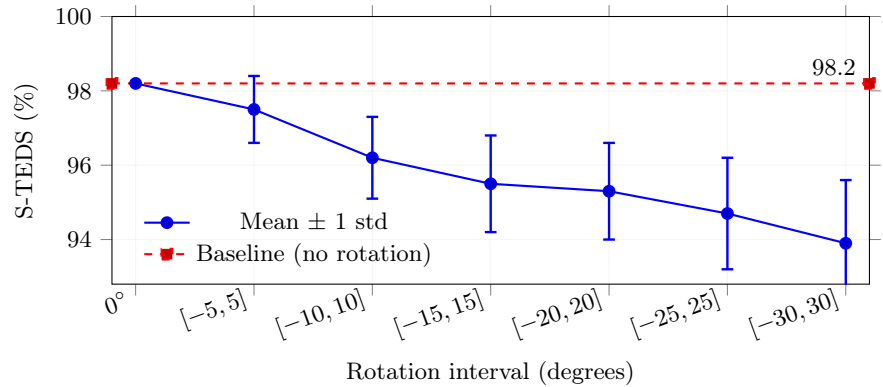

Figure~\ref{fig:curved_ftn_rotation} reports structure-only similarity (S-TEDS) on FinTabNet under random in-plane rotations. For each image, we sample a rotation angle uniformly from $[-\alpha,\alpha]$ and evaluate the rotated table; we report the mean and standard deviation across the evaluation set. Performance decreases smoothly as the rotation range increases, from the no-rotation baseline (98.2) to 97.5 for $\alpha{=}5^\circ$ and 93.9 for $\alpha{=}30^\circ$, indicating that the curved-separator formulation preserves strong structure recovery under moderate rotations but remains sensitive to larger geometric perturbations; qualitative examples are shown in Fig.~\ref{representatives_samples}.

\section{Ablation study}
\label{sec:ablation}

\subsection{Impact of the Tiny Recursive Module (TRM)}
\label{subsec:TRM_ablation}
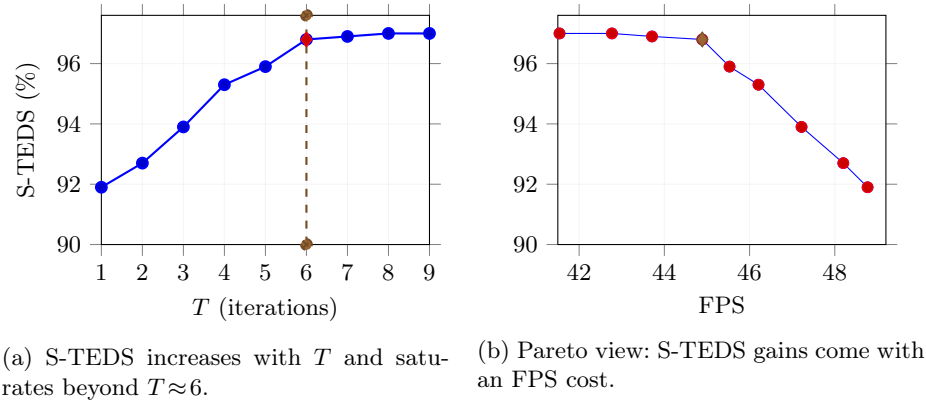
\begin{figure}[t]
  \centering
  \captionsetup{font=normalsize}
  \begin{subfigure}[t]{0.485\linewidth}
    \centering
    \vspace{0pt}
    \begin{tikzpicture}
      \begin{axis}[
        width=\linewidth,
        height=0.78\linewidth,
        xlabel={$T$ (iterations)},
        ylabel={S-TEDS (\%)},
        xmin=1, xmax=9,
        ymin=90, ymax=97.6,
        xtick={1,2,3,4,5,6,7,8,9},
        ytick={90,92,94,96},
        grid=major,
        major grid style={opacity=0.15},
        tick label style={font=\small},
        label style={font=\small},
        tick align=outside,
      ]
        \addplot+[mark=*, thick] coordinates {
          (1,91.9) (2,92.7) (3,93.9) (4,95.3) (5,95.9)
          (6,96.8) (7,96.9) (8,97.0) (9,97.0)
        };
        \addplot+[mark=diamond*, only marks, mark size=2.8pt] coordinates {(6,96.8)};
        \addplot+[dashed, thick] coordinates {(6,90) (6,97.6)};
        % \node[anchor=south west, font=\small] at (axis cs:6,97.15) {$T=6$};
      \end{axis}
    \end{tikzpicture}
    \caption{S-TEDS increases with $T$ and saturates beyond $T\!\approx\!6$.}
  \end{subfigure}\hfill
  \begin{subfigure}[t]{0.485\linewidth}
    \centering
    \vspace{0pt}
    \begin{tikzpicture}
      \begin{axis}[
        width=\linewidth,
        height=0.78\linewidth,
        xlabel={FPS},
        ylabel={},
        xmin=41.5, xmax=49.2,
        ymin=90, ymax=97.6,
        ytick={90,92,94,96},
        grid=major,
        major grid style={opacity=0.15},
        tick label style={font=\small},
        label style={font=\small},
        tick align=outside,
      ]
        \addplot+[thin] coordinates {
          (48.77,91.9) (48.20,92.7) (47.22,93.9) (46.21,95.3) (45.53,95.9)
          (44.89,96.8) (43.71,96.9) (42.77,97.0) (41.54,97.0)
        };
        \addplot+[only marks, mark=*, mark size=2.0pt] coordinates {
          (48.77,91.9) (48.20,92.7) (47.22,93.9) (46.21,95.3) (45.53,95.9)
          (44.89,96.8) (43.71,96.9) (42.77,97.0) (41.54,97.0)
        };
        % Highlight the chosen trade-off point.
        \addplot+[mark=diamond*, only marks, mark size=2.8pt] coordinates {(44.89,96.8)};
        % \node[anchor=south east, font=\small] at (axis cs:44.89,96.8) {$T=6$};
      \end{axis}
    \end{tikzpicture}
    \caption{Pareto view: S-TEDS gains come with an FPS cost.}
  \end{subfigure}

  \vspace{-1mm}
  \caption{\textbf{Effect of TRM iterations $T$ on PubTabNet.}
  (a) Accuracy improves rapidly for small $T$ and saturates after $T\approx 6$.
  (b) Throughput decreases with $T$, revealing an elbow around $T=6$ that provides a favorable accuracy--speed trade-off.}
  \label{fig:trm_ablation}
\end{figure}

FastTab refines the global latent representation $z$ through a small number of recursive refinement steps. To quantify the effect of this iterative refinement, we train FastTab with different numbers of TRM iterations $T$ and report S-TEDS on PubTabNet (validation). Fig.~\ref{fig:trm_ablation}(a) shows a clear accuracy--iteration trend: moving from $T{=}1$ to $T{=}4$ yields a strong absolute improvement (from $91.9\%$ to $95.3\%$, i.e., $+3.4$ points), indicating that a few refinement steps already correct a large fraction of global structural inconsistencies. Gains persist but become smaller beyond $T{=}4$: the improvement from $T{=}5$ to $T{=}6$ is $+0.9$ points ($95.9\%\rightarrow 96.8\%$), while additional iterations bring only marginal benefit ($\le 0.2$ points from $T{=}6$ to $T{=}9$). This saturation suggests that most recoverable long-range constraints (e.g., consistent separator placement and coherent spanning patterns) are already captured with a moderate recursion depth, and that further refinement mostly performs minor corrections.

The computational cost of refinement is captured in Fig.~\ref{fig:trm_ablation}(b). Increasing $T$ reduces throughput almost monotonically, from $48.8$ FPS at $T{=}1$ to $41.5$ FPS at $T{=}9$ (a relative drop of $\approx 15\%$), reflecting the additional TRM computations and the downstream dependence on refined global context. Importantly, the Pareto view highlights that the regime $T\ge 6$ is dominated: it incurs a non-negligible FPS loss for negligible S-TEDS gain. We therefore adopt $T{=}6$ as the default setting in the main experiments, as it lies near the elbow of the accuracy--speed curve and offers a good balance between structure quality and inference efficiency.

\subsection{Axis modeling}
\label{subsec:ablation_axis}

FastTab predicts row/column separators from a high-resolution feature map $\mathbf{F}\in\mathbb{R}^{d\times H_f\times W_f}$. The default design performs an \emph{axis-wise projection} followed by \emph{light sequence modeling}: we compress $\mathbf{F}$ into two 1D sequences
$\mathbf{S}^{(r)}=\mathrm{mean}_w(\mathbf{F})\in\mathbb{R}^{d\times H_f}$ and
$\mathbf{S}^{(c)}=\mathrm{mean}_h(\mathbf{F})\in\mathbb{R}^{d\times W_f}$, project them to a smaller embedding dimension $d_{\mathrm{seq}}$, and apply a tiny 1D Transformer along each axis. This ablation tests (i) whether early projection discards useful 2D cues for separator localization, and (ii) whether self-attention is necessary beyond cheaper local operators.

We keep the encoder, TRM, interval decoding (softmax over intervals + cumulative-sum boundaries), grid generator, and spanning head identical across all variants, and replace \emph{only} the two axis branches that output row/column interval logits. All models are trained with the same data, schedule, and losses. We report S-TEDS on validation and end-to-end FPS (image $\rightarrow$ HTML, including spanning inference), measured with the same batch size and hardware.

Table~\ref{tab:ablation_axis} compares four separator heads. In all cases, the head consumes the same feature map $\mathbf{F}$ and produces the same fixed-length interval logits $(R_{\max}{+}1)$ and $(C_{\max}{+}1)$ so that training and decoding remain unchanged.

\begin{itemize}
  \item \textbf{Axial + MLP (no context).}
  After projection, each position is processed independently by a pointwise MLP (equivalently, a $1\times 1$ Conv1D) with hidden size $h{=}256$. This baseline isolates the contribution of \emph{any} axis modeling beyond the encoder.

  \item \textbf{Axial + 1D Conv (local context).}
  We replace self-attention with a residual stack of $L{=}2$ Conv1D blocks (kernel $3$) at width $d_{\mathrm{seq}}{=}256$. This variant captures short-range smoothing while remaining strictly local.

  \item \textbf{Axial + Tiny 1D Transformer (default).}
  We use the FastTab axis module: a channel projection to $d_{\mathrm{seq}}{=}256$ followed by a $L{=}2$-layer 1D Transformer (4 heads, FFN dim 512) along each axis. This provides long-range interactions required for globally consistent separators.

  \item \textbf{2D head (no early projection).}
  We retain $\mathbf{F}$ in 2D and predict two separator heatmaps with a lightweight depthwise-separable CNN head ($3$ blocks, kernel $3$). The heatmaps are reduced to 1D signals only at the end by averaging along the orthogonal axis and then mapped to interval logits, thus isolating the effect of early axis projection while keeping the same decoding.
\end{itemize}

\begin{table}[t]
  \centering
  \small
  \caption{Axis modeling ablation on \textbf{PubTabNet} (validation). All variants share the same encoder, TRM, decoding, and spanning head; only the separator head differs. Reported FPS is end-to-end.}
  \label{tab:ablation_axis}
  \begingroup
  \setlength{\tabcolsep}{6pt}
  \renewcommand{\arraystretch}{0.95}
  \begin{tabular}{l c c}
    \toprule
    \textbf{Separator head}  & \textbf{S-TEDS (\%)} & \textbf{FPS} \\
    \midrule
    Axial + MLP          & 94.4 & 45.31 \\
    Axial + 1D Conv      & 95.7 & 45.05 \\
    Axial + 1D Transformer (FastTab) & 97.4 & 44.28 \\
    2D head (no projection) & 97.6 & 37.05 \\
    \bottomrule
  \end{tabular}
  \endgroup
\end{table}

Table~\ref{tab:ablation_axis} shows that contextual modeling along each axis is crucial: the Tiny 1D Transformer improves S-TEDS by $+1.7$ points over a local Conv1D baseline (95.7$\rightarrow$97.4) with a negligible FPS drop (45.05$\rightarrow$44.28). Retaining full 2D processing yields only a marginal additional gain (+0.2 S-TEDS) but is substantially slower (37.05 FPS), suggesting that early axis-wise projection is largely sufficient once encoder features are strong, and that the main benefit comes from long-range 1D interactions rather than preserving 2D structure throughout.

\section{Conclusion and Future Work}
\label{sec:conclusion}

We presented \textsc{FastTab}, a fast grid-centric TSR model that avoids autoregressive HTML decoding by combining a Tiny Recursive Module (TRM) for table-level reasoning with axial 1D Transformers for long-range row/column context. The model predicts counts, header rows, and separators to build a grid, then infers rowspan/colspan from ROI-aligned cell features, yielding a complete table structure.

Experiments on PubTabNet, FinTabNet, PubTables-1M, and SciTSR show that \textsc{FastTab} achieves competitive structure recovery with low-latency. Ablations confirm that TRM iterations improve S-TEDS up to a moderate depth and that axial self-attention substantially outperforms local axis baselines with minimal speed impact, while full 2D processing brings only marginal gains at a higher cost. We also observed robustness to anonymisation methods that preserve layout cues and demonstrated an extension to curved separators for improved tolerance to moderate geometric distortions.

Future work includes extending \textsc{FastTab} to multi-table pages by integrating detection/layout analysis, improving robustness to stronger rotations and perspective warps, and replacing fixed $(R_{\max},C_{\max})$ caps with adaptive or hierarchical grid decoding for very large tables.

\subsubsection{\discintname}
The authors have no competing interests to declare that are relevant to the content of this article.

\bibliographystyle{splncs04}

\end{document}